\documentclass[letterpaper, 10 pt, conference]{ieeeconf}  

\IEEEoverridecommandlockouts                              

\usepackage{xcolor}
\usepackage{url}
\usepackage{graphicx}

\title{\LARGE \bf
Judging by the Look: The Impact of Robot Gaze Strategies on Human Cooperation
}

\author{Di Fu$^{\dagger}$*, Fares Abawi$^{\dagger}$*, Erik Strahl$^{\dagger}$ and Stefan Wermter$^{\dagger}$
\thanks{*Di Fu and Fares Abawi contributed equally to this work.}
\thanks{$^{\dagger}$Di Fu, Fares Abawi, Erik Strahl, and Stefan Wermter are with Knowledge Technology, Department of Informatics,
        University of Hamburg, 22527 Hamburg, Germany
        {\tt\small \{di.fu, fares.abawi, erik.strahl, stefan.wermter\}@uni-hamburg.de}}%
}

\begin{document}

\maketitle
\thispagestyle{empty}
\pagestyle{empty}

\begin{abstract}

Human eye gaze plays an important role in delivering information, communicating intent, and understanding others' mental states. Previous research shows that a robot's gaze can also affect humans' decision-making and strategy during an interaction. However, limited studies have trained humanoid robots on gaze-based data in human-robot interaction scenarios. Considering gaze impacts the naturalness of social exchanges and alters the decision process of an observer, it should be regarded as a crucial component in human-robot interaction. To investigate the impact of robot gaze on humans, we propose an embodied neural model for performing human-like gaze shifts. This is achieved by extending a social attention model and training it on eye-tracking data, collected by watching humans playing a game. We will compare human behavioral performances in the presence of a robot adopting different gaze strategies in a human-human cooperation game.

\end{abstract}

\section{INTRODUCTION}

Eye contact plays an important role in interpersonal communication. A recent study shows eye contact can increase the synchronization between humans' brains~\cite{luft2022social}. Maintaining eye contact can deliver information, allowing for the inference of intent, and understanding others' mental states. Likewise, robot gaze can influence human decision-making during gameplay by lowering reaction time and increasing the cognitive effort~\cite{belkaid2021mutual}. The authors, however, design a robot gaze strategy following heuristic findings rather than simulating human eye movements, resulting in less realistic gaze shifts by the robot. Another study introduces a social gaze-control system to simulate visual human attention~\cite{zaraki2014designing}. However, there is no direct human-robot interaction in the experimental scenario. Moreover, the impact of robot gaze on human decision-making is not studied.

Overall, previous work suffers two crucial limitations: 1)~Gaze simulation is not based on eye-tracking data; 2)~Due to the disregard of human eye movements in designing robot gaze strategies, experiments do not examine the influence of such effects on humans. To address the aforementioned gaps, we propose two research goals: 1)~Simulate human gaze behaviors on robots based on eye-tracking data; 2)~Explore the influence of a robot's gaze interaction on the performances of humans in a cooperation game.

\section{INTERACTION SCENARIO, TASKS, AND PROCEDURE}
\begin{figure*}[thpb!]
      \centering
       \includegraphics[height=0.155\textheight]{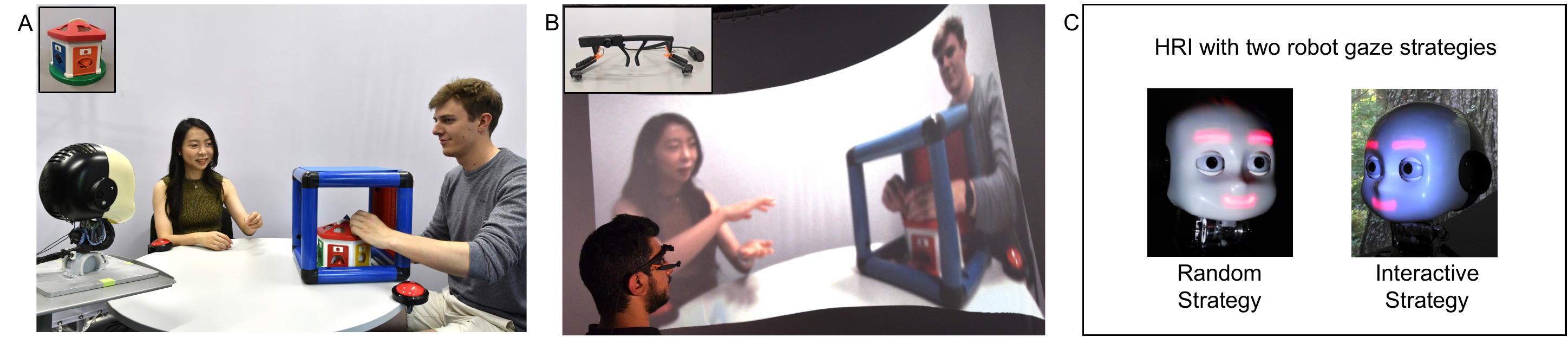}
       \caption{Three tasks comprising our experimental setup starting with \textit{A)} task 1 in which the robot remains static and records a pair of participants playing a cooperation game with a shape sorter (top left), followed by \textit{B)} task 2 where the head and eyes of a participant watching the gameplay are recorded using an eye tracker (top left). In \textit{C)} task 3 the robot performs movements based on either strategy while participants play the game described in task 1. }
       \label{fig_tasks}
   \end{figure*}

Our experiments will be performed in three separate tasks conducted in sequence as illustrated in Figure~\ref{fig_tasks}.
\subsection{Task 1: Human-human cooperation game with an inactive robot}

In task 1, participants will be randomly matched in pairs to play multiple rounds of a human-human cooperation game. We will use the iCub\footnote[1]{iCub: \url{https://icub.iit.it/}} robot in our experiments. The robot will display facial expressions in a Wizard-of-Oz setup without performing gaze or head movements. During the task, a pair of participants will sit around a table facing each other and play a game. The iCub robot will act as the instructor by asking one of the participants to place a particular shape in its corresponding hole on a shape sorter, e.g. \textit{``place the cylinder in the round hole"}. One participant will play the role of an actor, placing his or her hands in a box occluding the available objects and containers. The other participant will assume the role of a guide by helping the actor place the right object in the designated container through speech and gestures. After each round, the iCub robot will change its facial expressions. The participants will be asked to guess the intention behind the different facial expressions during the experiment. In doing so, the participants would distribute their attention between the task at hand and the iCub robot. The guiding participant will be asked to push a button once the round is completed to measure the response time. There will be 20 rounds in total. Each pair of participants will flip roles as actors and guides after every 5 rounds by switching seats. The 20-round response time will be summed up to be the total cooperation time for each pair. After finishing the cooperation game, the participants will fill in the Godspeed questionnaire~\cite{bartneck2008measuring} to rate their impression of the iCub robot. The entire game session will be recorded using the iCub's cameras and binaural microphones.

\subsection{Task 2: Social attention model training on human eye-tracking data}

In task 2, we will collect human eye-tracking data by recruiting participants to watch the videos recorded in task 1. We will then train a social attention model based on this eye-tracking data. The videos will be displayed on a large curved screen with stereo audio playback. The participants' eye movements will be recorded under the free-viewing condition while wearing a Pupil Core eye tracker\footnote[2]{Pupil Core: \url{https://pupil-labs.com/products/core/}}. The camera view in task 1 is from the iCub’s perspective, therefore, the robot is not visible in task 2. A saliency prediction model~\cite{AWW21} with crossmodal social cue integration will be trained on the collected human eye-tracking data. The model will be extended with a mapping scheme for associating perceivable stimuli within the auditory and visual fields. Such a model requires binaural perception to localize sound sources when visual stimuli do not lie within the visual field. Our model will, therefore, combine priority maps arriving from different modalities to form a master map for attending to elements with high conspicuity, regardless of their visibility.

\subsection{Task 3: Human-human cooperation game with human-robot joint and mutual gaze interaction}
The social attention model trained in task 2 will be mounted on the iCub robot. In this task, there will be two strategies guiding the robot's gaze behavior - interactive and random. The interactive strategy will follow the predictions of our social attention model trained on the eye gaze data acquired in task 2. This strategy will likely increase the robot's joint and mutual eye gaze with the participants. The random strategy model will, however, perform gaze shifts based on statistical assumptions regarding fixation time and saccadic movements. This model is agnostic to the stimuli, allowing for robot behavior similar to the typical human without perceiving the environment. Pairs of participants will be recruited to play the same game described in task 1, for a total of 40 rounds. As opposed to task 1, the iCub will perform gaze movements based on either strategy. The participants' response time will be recorded given their exposure to the iCub following the interactive and random gaze strategies for 20 rounds per strategy. The condition (strategy) ordering will be shuffled to avoid order bias. 

\section{DATA ANALYSES AND EXPECTED RESULTS}

The sum and the average reaction time for each pair of participants will be calculated. One-way ANOVA will be measured on these statistics for the three robot conditions (inactive/interactive/random) to detect the impact of different human-robot gaze interactions on human cooperation. Ideally, if the pairs remain unchanged between tasks 3~and~1, an independent t-test could be implemented by comparing the test-retest differences between the interactive and random groups. The inactive condition will be taken as a first test baseline. For the analyses of participants' ratings on the Godspeed questionnaire, one-way ANOVA will be measured for the sum of different dimensions across three groups.

We hypothesize that the social attention model can integrate multiple crossmodal cues and predict social saliency based on human data. Moreover, we expect that the interactive gaze strategy could increase human-human cooperation by decreasing the reaction time during gameplay compared to the other conditions. 

\section{CONCLUSIONS}

We propose to train a social attention model on human gaze behaviors in a crossmodal environment. The model will be deployed on the iCub robot as an embodiment platform. Human decision-making and cooperation behaviors will be studied under different robot gaze conditions to explore the impact of joint and mutual attention in human-human and human-robot interaction. Our work combines behavioral findings and computational modeling to bridge the gap between cognitive simulation and human-robot interaction.


\addtolength{\textheight}{-12cm}   





\section*{ACKNOWLEDGMENT}

The current work is supported and funded by the German Research Foundation DFG under project CML (TRR 169).



\bibliographystyle{IEEEtran}
\bibliography{root}

\end{document}